Lothar Lemnitzer, Laurent Romary, Andreas Witt

**Representing human and machine dictionaries in Markup languages (SGML, XML)**

## 1 Introduction

Mankind has been using, in a long history of preserving and storing knowledge, a wide range of media for this purpose. Over the centuries, clay tablets, papyri, parchment, paper and, most recently, all kinds of discs have been used to store and therefore preserve information that was, in its time, considered worthy of preservation. Reference works have always been an integral part of this development (cf. McArthur, 1988).

Today we are facing another media revolution, with a period of synchronous use of so called "analog" media – e.g. paper and print – and "digital media" –e.g. hard discs, CD-ROMs and other storage media for digital data.

Producers and publishers of reference works such as dictionaries must cope with this situation. Economic as well as other reasons demand that a) existing "paper and print" dictionaries be digitized for use on a desktop or over the internet – the so-called (retro-)digitization of existing print dictionaries – and b) new dictionaries be produced in a way that anticipates their use in all kinds of (old and new) devices.

Dictionaries can be seen, in a somewhat simplified view - as collections of entries. These entries are highly structured objects where the various text components bear neither the same semantic nor the same internal organization. In paper and print dictionaries, typographic (e.g. font weight) as well as non-typographic means (e.g. bracketing) are used to convey this structural information to the user. With the advent of new media and the digitization of dictionaries, a tendency to separate the logical structure and the layout characteristics of dictionary entries has taken place. One advantage of such a separation is that information traditionally expressed by typographic conventions becomes explicit both for humans and for computer programs. The common way of doing this is to markup the various parts of texts, i.e. to insert additional information in order to signify the semantics of these parts.

The media specific layout of the articles and their parts is added by means of style-sheets (see sect. 5.1).

In the following section, we will describe typical entry structures found in traditional paper and print dictionaries. This structure will be illustrated by means of two examples of traditional dictionaries that have recently been digitized.

In section 3, we will briefly  introduce markup languages which are suitable for modelling the tree-like structure of dictionary entries and for describing the semantics of the individual components. Document grammars, also called schemata, i.e. specialized mechanisms to constraint the structure of annotations, will be introduced as well.

Two document models for dictionary articles are currently in use and can be seen as (quasi-)standards for the modeling of print articles: a) the Guidelines of the Text Encoding Initiative (TEI, Sperberg-McQueen, C.M./Burnard L. 2004), in particular chapter 12, and b) the ISO 1951 standard for printed dictionaries.

While the TEI Guidelines have a long tradition, enjoy a high reputation in the field of academic dictionary projects and are supported by a large community with examples of good practice, schemas, tools etc., this is not the case with ISO 1951. As we will explain in section 5, this standard still suffers from conceptual shortcomings. It might, however, play more of a role in dictionary publishing in the future if the conceptual shortcomings are amended.

The respective standards are described in section 4.

Most recently, another ISO standard, called "Lexical Markup Framework", was released. While this framework is clearly oriented towards the presentation of dictionaries for natural language processing, it can and should also be related to the TEI and ISO 1951 standards. Therefore, LMF is briefly described in section 4 as well.

Section 5 is devoted to transformation tools and scenarios. Once a dictionary article is modeled as a tree-like-structure, and the individual elements are given an explicit semantics, these articles can be transformed in various ways to either a) transform them into other trees by re-arranging the parts or deleting some elements or b) derive a media-specific representation format from the source. We will describe XSLT as one special purpose programming language to specify and perform such transformations. Furthermore, we will use a simple mapping from TEI encoding to ISO 1951 encoding as one example.

Markup languages and schemata in particular can also be used to control the dictionary compilation process. The use of a) well defined article structures and b) pre-defined vocabularies can be enforced by the use of customised schemata. We will explain how this works, using again the examples introduced in section 2.

This article closes with some concluding remarks and a bibliography.

## 2 The tree-like structure of dictionaries and dictionary entries

At first sight, dictionary entries appear to the reader's eye as a sequence of words and symbols. This holds for most printed as well as electronic dictionaries (see the examples in figure 1 and 2, excerpted from printed dictionaries).

At second sight, one might detect a non-linear structure underlying this sequence. This structure resembles the mathematical model of a tree, i.e. an ordered, directed, and acyclic graph.

There are typographical as well as non-typographical signals that indicate where a segment of a dictionary article starts and where it ends. Indeed, it is essential to understand this underlying hierarchical structure in order to use a dictionary article efficiently. Learned users of dictionary articles, such as given in example 1 and 2, can easily find their way to the information that they need in a certain situation.

More formally, the typical dictionary article consists of a fixed order sequence of segments, where each segment conveys a unique piece of information about the object that is described in an article – typically a lexical entity of some kind (morpheme, word, phrase etc.).

Another feature of the dictionary entry structure is not so obvious to the user. Most dictionary articles are organized hierarchically. An entry may consist of several sub-parts, each of which conveys information about different homographs of a headword (alternatively, homographs could be represented as two or more independent entries). It may consist of a description of the form, function and grammatical features of the linguistic sign as well as of numerous senses, each of which may in turn be composed of some sub-senses etc.

On this more abstract level, we can model a dictionary entry as a tree with a root (which represents the full entry), intermediate levels (e.g. forms section, semantic section) and the individual segments at the leaves.

In the following we introduce (fragments of) two sample entries that will be referred to in the remainder of this article. The print dictionaries from which these articles are excerpted have been digitized, encoded and annotated using XML in conformance with the TEI Guidelines (see below).

---

**Bahn-** ...- **hof**, der, *Halle, Gebäude am Halteplatz von Eisenbahnzügen*: am B. sein; jmdn. Am B. erwarten, vom B. abholen, zum B. bringen; auf welchem B. kommt er an?; wie weit ist es bis zum B.?; der Zug rollte aus dem B.; im Gedränge des Bahnhofes; Neupräg. salopp ich verstehe immer nur B. (*ich verstehe gar nichts*); Neupräg. großer B. *festlicher Empfang*: der berühmte Gast wurde mit großem B. empfangen; von einem großen B. absehen

---

Figure 1: Entry for the headword *Bahnhof* ('railway station') excerpted from the Wörterbuch der deutschen Gegenwartssprache (=WDG), 6. Vol., ed. by Ruth Klappenbach und Wolfgang Steinitz, Berlin 1962-1977, online at: www.dwds.de. The entry is slightly abridged to make it easier to read.

---

**Ski** m., seit Anfang 20 Jh. meist *Schi,* ‚Scheeschuh', Übernahme (Anfang 19. Jh.) von gleichbed. norw. s*ki*, aus anord. *sk..* ‚Scheit, Schneeschuh'; s. das im Dt. etymologisch entsprechende *Scheit.*

---

Fig 2: an entry from: Wolfgang Pfeifer: Etymologisches Wörterbuch des Deutschen. 2. Aufl. Akademie Verlag 1993. Online available at: <u>www.dwds.de</u>)

While digitizing these articles, as well as the rest of both dictionaries, a XML encoded version of these entries has been produced. The markup used to annotate these dictionaries is an adaptation of the TEI guidelines for printed dictionaries. The results are shown in fig. 5 and 6. The abstract entry
structure has been specified by a RELAX NG schema (see section
6 below).

  The entries consist of different parts. Both entries have a
form section and a sense section. The form section contains
information about the spelling of the headword (i.e. Bahnhof,
Ski), including variants (e.g. Schi), its part of speech as
well as gender (e.g. 'm.') and additional grammatical
information. The sense section groups all readings of the
headword in a hierarchical manner i.e. there are main senses
and sub-senses. The meaning description of ex. 1 makes use of
a definition (e.g. '<u>Gebäude am Halteplatz von Eisenbahnzügen',</u>
<u>tr. building located at regular stop points of railway</u>
<u>trains</u>), usage examples (e.g. vom Bahnhof abholen, tr. to
fetch from the railway station) and citations.

  In the second example, the entry also contains an etymology
section which informs the reader, in textual style, about the
history of the described lexical unit, e.g. Norwegian and Old-
Nordic.

  The description above makes the tree-like structure of the
dictionary entries obvious. However, at one point the article
deviates from the strict hierarchical and acyclic model: there
is a cross-reference in the etymology section of the second
example (i.e. 's. das im Dt. etymologisch entsprechende
*Scheit*', tr. see the etymologically related German word
*Scheit*), which points to another article in the same
dictionary. Standards for dictionary (article) encoding take
care of the so called "meso-structure" or "cross-reference
structure" features of  dictionaries by providing reference
mechanisms to uniquely identifiable elements (i.e. articles or
even particular parts of articles).

  As we have seen, dictionary entries are highly structured
objects where the various text components bear neither the
same semantic nor the same internal organization. This holds
true both within individual dictionaries and across categories
of dictionaries. At the same time, representing a dictionary
in digital format should preserve the linearity of its
content, which is essential for human legibility. We thus need
a *markup language* that copes with both aspects, while enabling adequate processability by machines.

## 3 SGML and XML

The eXtensible Markup Language (XML) was designed by the
World Wide Web Consortium (W3C) at the end of the 1990ies to
fulfill exactly the abovementioned needs in the context of
web-based applications. It was elaborated as a simplification
of the older Standard Generalized Markup Language (SGML), an
ISO standard for structured documents published in 1988 (ISO
8879), as well as a generalization of mechanisms already
present in the HTML language for presentation of online
information.

XML basically introduces a syntax for marking up sections of
a document by means of *opening* and *closing tags*, which form,
together with the enclosed content what is called an *element*.
For instance, if one wants to markup in a dictionary entry
that the string "f" should be tagged as belonging to the
category "gen" (for gender), the following example is a so-
called *well-formed* XML fragment:

```
<gen>f</gen>
```

In this example, "<gen>" is the opening tag, "</gen>" the
closing tag and "f" the actual (here textual) content of the
element. Note that there can also be empty elements, for
instance "<pb/>" for indicating page breaks within a document,
and elements with mixed contents, i.e. elements which contain
child elements as well as textual content on the same level.

Furthermore, XML allows such bracketed information to be
hierarchically nested in order to form complex, marked up
structures. For instance, a structure grouping information
about a meaning (<sense>) in a dictionary entry, based on a
definition (<def>), and a cited example (<cit>, together with
<q> for the cited text and <bibl> for the corresponding
bibliographical source) may be sketched as follows (for
reasons of simplicity, the content of the elements is
represented by placeholders, see fig. 5 for a full-fledged
example):

```
<sense>
    <def>...</def>
```

```
    <cit>
         <q>...</q>
         <bibl>...</bibl>
    </cit>
</sense>
```

Finally, the XML syntax allows one to add further information to a tagged section by means of *attributes*, which appear within the opening tag of an XML element. Such attributes may either be specific to the encoding scheme, for instance to number the meanings in a dictionary entry as in:

```
<sense n="1">...</sense>
```

When defined directly by the W3C consortium generic to all XML applications (indicated by means of a prefix to the element and attribute names; in the case described here, the prefix 'xml:' is used conventionally), as is the case for the indication of the working language as follows:

```
<q xml:lang="fr"> Le langage des fleurs et des choses muettes!</q>
```

To provide unique identifier for any XML element (for instance entries in a dictionary) within a document, as follows:

```
<entry xml:id="ent2352"> ...</entry>
```

If "xml:id" is of the data type "identifier" (ID), the software which processes the document has to ensure that the same value is assigned to this attribute but once in a document. In other words: the value of "id" has to be unique.

The important underlying characteristic of XML is that all marked up content forms a strict hierarchy, which in turn, from a computational point of view may be represented as a tree structure. For instance, the <sense> example above may be outlined as in figure 3, to delineate its underlying tree structure.

As is the case in the dictionary example 'ski', which contains a link to another dictionary entry, it is also possible to deviate from the strict tree-like structure in XML documents. The attributes "id" and "idref" (i.e. reference to an id) are a way to handle this issue of cross-references.

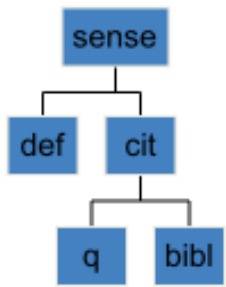

Figure 3: tree representation of the above XML fragment

Contrary to the situation in relational databases, the actual possible structures for such documents are not strictly limited to one single tree, but may vary according to design choices, for instance when one wants to iterate over senses in a dictionary, or make senses occur recursively within other senses. This is why the corresponding documents are often referred to as *semi-structured* in the literature (cf. Buneman, 1997).

In order to control which structures are allowed in the context of a given application, it is necessary to design a document grammar or *schema*, which will state which elements may occur under which conditions in a document. Indeed, there are several technologies to do so which have been either defined specifically by the W3C consortium (DTD or W3C XML schemas) or within other standardization bodies such as ISO (e.g. Relax NG schemas). Such schema (or document grammar) languages are based on the same principles, which can be summarized as follows:

- They restrict the number of so-called *valid* documents by stating explicitly which elements, attributes and data types are allowed;

- They define the *content model* of each element by indicating which element and text nodes are allowed as its *children* and in which order and how many times they may appear;

- For attributes, as well as for leaf elements (i.e. elements not containing further elements), they provide

mechanisms to constrain  content to specific data types (e.g. number, date, URI) or lists of possible values.

Different schema languages allow different types of constraints to be expressed.
In the above examples, e.g. 'Ski' and 'Bahnhof', it would be desirable to express the following constraint: 'a gender element is allowed only if the value of the part of speech element or attribute equals "noun"'. However, in most schema languages it is not possible to model interactions between parts of a document when the value of one element or attribute puts constraints on the structure or the values of other elements. Nevertheless, there are schema languages which can impose such constraints and therefore restrict the set of well-formed (entry) structures, e.g. Schematron (ISO 19757-3)

The main issue at hand when designing an XML schema is to identify which information is relevant for, on the one hand, a proper identification of the content to be represented, and, on the other hand, to match the usage that one wants to achieve with the information. For instance, if one wants to present information as web pages only, with no intention whatsoever to provide a stable and reusable representation, it may be tempting to just adopt the basic HTML syntax. For instance, one would thus indicate that part of speech information is to be presented in bold face by using the following HTML representation:

```
<b>noun</b>
```

On the contrary, if more than one possible usage is anticipated for dictionary content, and one is  determined to give, above all,  a dedicated semantic to tags to be used, a more explicit syntax (like the one of the TEI, see below) is to be adopted. This would make the previous example look like:

```
<pos>noun</pos>
```

where "<pos>" refers to the well-known linguistic concept of "part of speech".

The idea here is to abstract away from visual representations of dictionary entries, given that XML has available to it specific technical environments (CSS or XSLT and XSL-FO) dedicated to the actual presentation or transformation of semi-structured documents.

## 4 Standardized approaches for XML-based representations of dictionary-entries

   The most relevant document model for the digital representation of dictionary articles is currently the dictionary component of the modular annotation scheme defined by the Text Encoding Initiative (TEI, Sperberg-McQueen, C.M./Burnard 2004). As an alternative, another standard dealing with the representation format for printed dictionaries, ISO standard 1951, has emerged. Moreover, in 2008 ISO published the Lexical Markup Framework (LMF) as "a common model for the creation and use of electronic lexical resources" (ISO 24613). While LMF in general is described elsewhere in this volume (ch. 99) this section begins with a short presentation of LMF's Machine readable dictionary extension.

### 4.1 Lexical Markup Framework (LMF)

   The ISO-standard LMF defines a meta-format for lexical data. LMS's core package explains the basic hierarchy of information in a lexical entry. Additionally, LMF defines, inter alia, an extension for the representation of machine readable dictionaries.

   Since LMF defines a meta-model for lexical entries, it provides an abstract representation format for lexical information. LMF's main application area is NLP dictionaries but the ISO standard also defines an extension for machine readable dictionaries. Fig. 4 shows an LMF representation of the example entry "Ski" that makes use of the MDR extension.

```xml
<LexicalEntry>
 <Lemma id="l1">
  <FormRepresentation>
      <feat orthographyName="GermanVariantD"/>
      <feat writtenForm="Ski"/>
  </FormRepresentation>
  <FormRepresentation>
      <feat orthographyName="GermanVariantB"/>
      <feat writtenForm="Schi"/>
  </FormRepresentation>
 </Lemma>
 <Equivalent>
  <feat lang="German"/>
  <feat writtenForm="Schneeschuh"/>
 </Equivalent>
 <etymology>
```

```
  <etymon id="l2">
     <form>
      <orth xml:lang="norwegian">ski</orth>
      <pos>commonNoun</pos>
     </form>
     <sense>
      <gloss>device for sliding on snow</gloss>
      <note>aus anord. sk.. ,Scheit,
      Schneeschuh'; s. das im Dt. etymologisch
      entsprechende Scheit.</note>
     </sense>
   </etymon>
     <etymologicalLink source="l2" target="l1">
       <etymologicalClass>loan word
       </etymologicalClass>
      </etymologicalLink>
 </etymology>
</LexicalEntry>
```

Figure 4: The dictionary entry presented in fig. 2 encoded using the LMF

The given XML representation is a linearization of a more general model, in LMF normally visualized as a diagram in the Unified Modeling Language (UML) format. The lemma 'Ski' is given with its two alternative spellings and with the 'equivalent' lemma that is included in the original dictionary entry. Moreover, the example provides ample etymological information. The treatment of the latter within LMF is described in more detail by Susanne Alt (2006).

## 4.2. TEI specification for printed dictionaries

The Text Encoding Initiative (TEI; www.tei-c.org) is the most prominent international endeavour in the humanities, and probably beyond, to provide a set of reference guidelines for the representation of digital textual information. Initiated in 1987 as a forum of major text archives worldwide, it has been an early adopter of SGML, and in turn XML, and has issued five editions of its guidelines so far, providing more than 500 elements for representing prose, poetry, drama, manuscript and of course, dictionary content. The TEI has evolved step by step to become a real infrastructure for specifying and customizing textual formats, while allowing a quick entry into its technology (Romary 2009). It is based on a specification language (ODD – One Document Does it all), from which is generated, on the one hand, the full textual documentation, and on the other hand, the actual formal specification in several schema languages (DTD, Relax NG, W3C).

One important mechanism available in the TEI infrastructure is a class management system allowing elements to be grouped together when they bear either a joint semantics or when they occur in similar structural contexts. Classes are essential for providing an abstract entry point (e.g. all elements providing grammatical information in a dictionary entry) for the specification of a given construct, onto which one can easily customize a specialist profile (e.g. for my own dictionary, I just need *part of speech* and *gender*).

The "print dictionary" chapter of the TEI guidelines has been designed, right from the beginning (Ide & Véronis, 1995) as a generic model that could, in turn, be customized to deal with the variety of form and structure that print dictionaries or born digital machine readable dictionaries may take. It has also been a compromise between providing a highly structured format for controlling dictionary content, and accounting for the many and varied permutations and combinations that surface forms can take, especially in older dictionaries. This is why all elements belonging to the TEI dictionary chapter may occur within two main constructs:

- An <entryFree> element, where each component is tagged independently from one another in the order they appear in the printed text;
- An <entry> element, which provides a highly structured organisation of content, somehow closer to a database-like organisation.

As an example, the freely organised representation of the beginning of our "Bahnhof" example would appear as follows:

```
<entryFree><orth>Bahn- ...- hof</orth>, <gen>der</gen>, <def>Halle,
Gebäude am Halteplatz von Eisenbahnzügen</def>: <q>am B. sein</q>; <q>jmdn.
Am B. erwarten</q>...</entryFree>
```

The fully structured representation is depicted in figure 5.

```
<entry xml:id="E_b_437">
<form type="headword">
  <orth extent="suffix">-hof</orth>
  <gramGrp>
    <pos value="N"/>
    <gram type="determiner">der</gram>
</gramGrp>
</form>
<sense xml:id="S_b_234" level="0">
  <def xml:id="N_b_140">Halle, Gebäude am Halteplatz von
Eisenbahnzügen</def>
  <cit type="example">
    <quote>am B. sein</quote>
  </cit>
  <cit type="example">
    <quote>jmdn. am B. erwarten, vom B. abholen, zum B. bringen</quote>
  </cit>
  <cit type="example">
    <quote>auf welchem B. kommt er an?</quote>
  </cit>
  <cit type="example">
    <quote>wie weit ist es bis zum B.?</quote>
  </cit>
  <cit type="example">
    <quote>der Zug rollte aus dem B.</quote>
  </cit>
  <cit type="example">
    <quote>im Gedränge des Bahnhofes</quote>
  </cit>
<cit type="example">
  <usg type="reg">salopp</usg>
  <quote>ich verstehe immer nur B.</quote>
  <quote type="paraphrase">ich verstehe gar nichts</quote>
</cit>
</sense>
</entry>
```

Figure 5: The dictionary entry presented in fig. 1 encoded
using TEI

   One of the advantages of having these two possibilities at
hand is that it is possible to use these in a row, in the
context of a dictionary encoding workflow. Thus, it is
possible to see <entryFree> as an element that will allow an
initial surface markup of content preceding a second more
elaborate encoding, which will carve out the underlying
structure from the same content.

   We can now present the major characteristics of the TEI
dictionary principles in more detail. As mentioned in chapter
93, the TEI is conformant to the semasiological view of
lexical structures and actually maps rather precisely onto the LMF model, which has been introduced in the preceding section. This view is reflected in the TEI model through the provision of two main constructs:

- A <form> element, grouping together all descriptive elements (phonetic, orthographic, grammatical) of a surface realisation of the dictionary entry;
- A <sense> element, which organises, both sequentially and hierarchically, the various meanings associated to the entry

This high-level distinction between information about the formal aspects of the described linguistic sign and the content or meaning aspects nicely reflects usual practice when producing lexical descriptions in the form of a dictionary entry. A typical entry in a semasiological dictionary starts with an account of the formal features of the lexical sign, which are shared by all its senses.

In the case that the formal and / or etymological features of a linguistic sign vary even if the headword is the same, the TEI provides the <hom>-element to split an entry into two or more homographs.

The description of the individual senses or readings follows. Within a sense description, of course a part of the form description can be further specified or constrained. For example, a lexical unit which can in general be used in the plural form might not be used in a particular sense as such.

The <form>-element can be used to group the following types of information: spelling of a word, including alternative spellings (<orth>-element, see *Ski/Schi* in the entry example in fig. 1 and 6), other information related to the use of the word in written discourse, e.g, syllabification; pronunciation(s) of a word (<pron>-element) and other information  related to the use of the word in spoken discourse, e.g. word stress; grammatical information, e.g. part of speech, gender, inflection, to mention only a few prominent examples. Each piece of information can be further qualified with a "usage"-marker (the <usg>-element). The scope of a description can thus be constrained to a certain regional variety or a certain diachronic stage of the described language. Grammatical features of the described linguistic sign, e.g. part of speech, inflections, subcategorization, can be grouped into a grammatical-group structure (the <gramGrp>-element). Iteration as well as nesting of the <gramGrp>-elements allow the dictionary writer to group features which are functionally related so that the change of one feature implies the change of the other feature. For example, take a look at the gender and inflection of the German lexical unit "Gischt" (engl. "spin drift"). The gender varies without a corresponding change in the meaning of the word. The inflection of the word depends on the gender, and it changes accordingly. For both inflectional paradigms there is a restriction on the use of the plural. This can be modelled, using the TEI specification, as follows:

```
<form type="headword">
<orth extent="full">Gischt</orth>
<gramGrp>
<pos value="N"/>
<gramGrp>
<gram type="determiner">der</gram>
<gram type="genitive">-es</gram>
<gram type="plural">-e</gram>

</gramGrp>
<usg type="plev">auch</usg>
<gramGrp>
<gram type="determiner">die</gram>
<gram type="genitive">-</gram>
<gram type="plural">-e</gram>
</gramGrp>
<gram type="singular-preferred">Pl. ungebräuchl.</gram>
</gramGrp>
</form>
```

The <sense>-element assumes a similar function, i.e. grouping of elements that are related and together describe a particular sense (or reading) of the lexical unit. Prominent information types are definitions (<def>-element), citations (<cit>-elements> and usage examples, often taken from corpora (<example>-element).

In fig. 5 we see an entry with one sense and one sub-sense. The sub-sense is embedded in the main sense. The sub-sense is further qualified by a usage-marker indicating the register of the phrase ("salopp" ~ "colloquial"). Definitions as well as examples are given. Note that the example which is used to illustrate the subsense is itself accompanied by a paraphrase. The latter explains the meaning of the phrase.

In fig. 6, a TEI-encoding of the second dictionary introduced in section 2 is presented.

```xml
<entry xml:id="E_S_646">
 <form type="headword">
  <orth extent="full" rend="sep:comma">Ski</orth>
    <usg type="time">seit Anfang 20. Jh. meist</usg>
  <orth extent="full">Schi</orth>
  <gramGrp>
    <pos value="N"/>
    <gen value="masculine">m.</gen>
  </gramGrp>
 </form>
 <sense xml:id="S_S_646" level="0">
  <def>Schneeschuh</def>
 </sense>
 <etym xml:id="W_S_646">
  Übernahme (Anfang 19. Jh.) von gleichbed. norw. ski, aus anord. skīð
<def>Scheit, Schneeschuh</def> s. das im Dt. etymologisch entsprechende
<ref type="dict" target="E_S_165">Scheit</ref>
 </etym>
</entry>
```

Figure 6: The dictionary entry presented in fig. 1 encoded using the TEI

## 4.3. ISO 1951

ISO standard 1951 was initially designed in the seventies[1] to cope with the variety of codes used in printed dictionaries. It evolved in the nineties[2] to deal with layout aspects of dictionaries more widely. Following the natural trend to address the electronic encoding of print dictionaries, in the latest release (in 2007), an attempt was made to put a more comprehensive model for these together. Introducing explicitly in its scope the notion of representation of information, ISO 1951:2007 states: "it specifies a formal generic structure independent of the publishing media". Even if ISO 1951 does not actually provide a useful encoding scheme for the representation of print dictionaries, its underlying kinship with the TEI makes it an interesting example to explore, both in order to identify its weaknesses in providing a real generic model for dictionary representation and also to observe some of its design choices at meso- and micro-structural levels. ISO 1951 is also a good reference point for eliciting transformation mechanisms, as we will see in section 5.2.

---

[1] ISO 1951:1973 Lexicographical symbols particularly for use in classified defining vocabularies
[2] ISO 1951:1997 Lexicographical symbols and typographical conventions for use in terminography

Figures 7 and 8 present an encoding of the entry for 'Bahnhof' and 'Ski' ( introduced previously in this chapter), which is based on an inferred XML representation made by the authors of this chapter[3]. The actual data categories exemplified here represent only a small fraction of the actual set described in ISO 1951, which covers all major information units needed for dictionary encoding. The examples provided here also allow one to perceive the general principles underlying the organisation of dictionary entries according to ISO 1951 principles.

```xml
<DictionaryEntry>
        <HeadwordCtn>
            <Headword>Bahn</Headword>
            <Headword>-hof</Headword>
            <PartOfSpeech value="N"/>
            <Note type="linguisticNote">der</Note>
        </HeadwordCtn>
        <SenseGrp>
            <Definition>Halle, Gebäude am Halteplatz von
Eisenbahnzügen</Definition>
            <Example>am B. sein</Example>
            ...
                <ExampleCtn>
                    <Register>salopp</Register>

                    <Example>ich verstehe immer nur B.</Example>
                    <Gloss>ich verstehe gar nichts</Gloss>
```

```
            </ExampleCtn>
          </SenseGrp>
      </DictionaryEntry>
```

Figure 7: The dictionary entry presented in fig. 1 encoded using the ISO 1951

```xml
<DictionaryEntry>
        <HeadwordCtn>
            <Headword>Ski</Headword>
            <TemporalUsage>seit Anfang 20. Jh. meist</TemporalUsage>
            <Headword>Schi</Headword>
            <PartOfSpeech value="N"/>
            <GrammaticalGender>m.</GrammaticalGender>
        </HeadwordCtn>
        <SenseGrp>
            <Definition>Schneeschuh</Definition>
        </SenseGrp>
        <Etymology> Übernahme (Anfang 19. Jh.) von gleichbed. norw. ski,
aus anord. sk•ð
            <Definition>Scheit, Schneeschuh</Definition> s. das im Dt.
etymologisch
            entsprechende <SeeAlso>
                <Ptr href="E_S_165"/>Scheit
            </SeeAlso>
        </Etymology>
    </DictionaryEntry>
```

Figure 8: The dictionary entry presented in fig. 2 encoded using the ISO 1951

The ISO 1951 model is indeed based upon three generic mechanisms (called "compositional elements") to organise information within dictionary entries. These are:

- Containers, which allow one to refine a given data category (e.g. example) with complementary information (e.g. register, grammatical constraint) to contextualise the intended representation. In the example provided in figure 7, gloss information is given for the sentence "Ich verstehe immer nur B.";

- Blocks, which provide a means to further decompose a given representational structure (e.g. a sense) into more precise sub-components (e.g. sense variations according to register information);

- Groups, which correspond to information compounds within a dictionary entry. There are indeed only two groups in the actual ISO 1951 specification, namely, SenseGroup and HomographGroup.

  Containers are actually an essential mechanism for
information modelling in general. Indeed, this mechanism
allows one to consider in a uniform manner all constructs
related to a specific data category, allowing the occurrence
of the category itself (Example) or its corresponding
container (ExampleCtn) in exactly the same contexts.

  On the contrary, blocks and groups (to a lesser extent) are
of less utility in the context of semi-structural information
design because they  artificially prevent the actual
recursivity of the corresponding construct. For instance, it
is important in a dictionary structure, as elicited in the
LMF, to account for the recursive nature of senses. In this
respect, elements in ISO 1951 such as TranslationBlock or
HomographGroup seem to be superfluous with regards to
SenseGroup, which could indeed be used as the unique (and
recursive) structuring element for senses. This is in
particular demonstrated by the fact that many occurrences of
TranslationBlock in ISO 1951 appear as a unique child of a
SenseGroup.

  As a whole, ISO 1951 suffers from an incomplete design which
makes it hardly usable in concrete applications. Moreover, the
need for such a standard is not obvious considering that the
TEI has, for some twenty years, provided a well-maintained
framework for representing print dictionaries that many
electronic dictionary projects have actually implemented. We
argue that the next revision of the standard should integrate
the TEI tagset as the reference for implementing the proposed
model.

## 5 Processing XML-annotated dictionaries
### 5.1. XSLT Style-sheets

  The World Wide Web Consortium has provided an XML-based
"Stylesheet and Transformation Language" (cf. Kay, 2007). XSLT
is a declarative language, which is used to transform XML data
into other (semi-structured) documents. According to the W3C
recommendation, "a transformation expressed in XSLT describes
rules for transforming zero or more source trees into one or
more result trees". A transformation is achieved by a set of
template rules. These rules associate a pattern, which matches nodes in the source tree (i.e. elements, attributes, textual nodes etc.) with a "sequence constructor". In other words, XSLT relies on a formalism which addresses elements in a tree by the path which leads from the rule of the tree (or any other node in that tree) to this node. For this purpose XSLT builds on XPath, a formal language which has also been provided by W3C (cf. Berglund et al., 2007). Nodes and sets of nodes are manipulated by sequence constructors. With such constructors, nodes can be, amongst other things: a) moved from one place in the source tree to another place in the target tree; b) deleted from the target tree. Additionally, new nodes can be inserted into the target tree.

Thus, XSLT transformations can be used to a)convert information in a lexical database to media-specific representation formats. To give a real-world example of this, let us consider mobile devices for which the transfer of data is expensive or the display screen small. A publisher could decide to reduce an article to some essential pieces of information, e.g. grammar and definition(s). Other information of a more illustrative kind such as citations could be deleted from the target tree.

On web pages, complex articles could be presented in a compressed version and unfolded by the user who wants to see and consult more comprehensive articles. This has indeed been done on the website which features the "Wörterbuch der deutschen Gegenwartssprache" ([www.dwds.de](http://www.dwds.de)); b) apply text compression techniques used in paper and print dictionaries, where space is an expensive asset, but not in digital media, where nearly infinite storage space is available. For example, in a printed dictionary, inflectional information for nouns might be presented by the inflectional suffixes for the genitive singular and nominative plural. This has been done for decades now in paper dictionaries of the German language (e.g. *Bahnhof <-es, -höfe>*). In an electronic version of the dictionary the whole paradigm, comprising of eight paradigmatic forms, could be presented in tabular form; c) filter certain parts of an article, e.g to provide data in response to a user request; d) convert information between encoding standards. A case of this kind is presented in the following section.

## 5.2. XSLT Style-sheets for converting TEI to ISO 1951

This section focuses on illustrating the transformation
possibilities offered by XSLT in the specific case of
transforming TEI encoding structures into ISO 1951
representations. Although not exhaustive, the transformation
examples presented here cover most of the situations that can
be encountered when doing similar processing on, say TEI based
structures, either for filtering (as result to a query),
transformation (to any other format), or presentation (e.g. in
XHTML) purposes.

The rules (or templates in XSLT words) presented here are
intended to be used within an XSLT stylesheet canvas for
taking TEI compliant entries as input and generating ISO 1951
as output. To this end, the root element of the stylesheet
will appear as follows:

```xml
<xsl:stylesheet xmlns:xsl="http://www.w3.org/1999/XSL/Transform"
  xmlns:tei="http://www.tei-c.org/ns/1.0" version="1.0">
  <xsl:output encoding="UTF-8" method="xml"/>
…
</xsl:stylesheet>
```

where, beside the namespace declaration needed for the XSLT
instructions proper, the TEI namespace is declared (prefix:
tei) to map onto all encountered TEI elements. We can also
note that the first child element of <xsl:stylesheet>
(xsl:output) states that the output result will be XML data
with characters encoded in the UTF-8 (8-bit framed Unicode
encoding).

In the simplest case, an XSLT transformation will take a
simple element, usually corresponding to an elementary data
category, and isomorphically bring its content into the
equivalent (in our ISO 1951 transformation use case) or
associated (when presenting the content in XHTML for instance)
construct. For example:

```xml
<xsl:template match="tei:gen">
    <GrammaticalGender>
        <xsl:apply-templates/>
    </GrammaticalGender>
</xsl:template>
```

transforms the <gen> element from the TEI namespace into an ISO 1951 <GrammaticalGender> element.

The same type of transformations not only apply at the level of elementary data categories, but also, in a very simple manner, for mesoscopic objects as in the following example transforming the generic (and recursive) <sense> element in the TEI into the group <SenseGrp>:

```
<xsl:template match="tei:sense">
    <SenseGrp>
        <xsl:apply-templates/>
    </SenseGrp>
</xsl:template>
```

As is the case in any recursive transformation, and can already be anticipated here, such rules presuppose that all transformations occurring before or after the actual firing of the rule are adequately described.

Some slightly more elaborated rules may be needed when data categories between the two models to be mapped do not exactly match, i.e. the underlying representation models do not have the same level of granularity. In such cases, specific tests (respectively, constraint generations) have to be implemented in the corresponding templates.

These two cases actually occur in our TEI to ISO 1951 example. Typically, both models do not account for the indication of headword in a dictionary in the same manner. The TEI introduces a generic <form> element that can, through the appropriate use of the 'type' attribute, be further constrained to indicate whether the actual object being represented is the headword or an inflected form. On the contrary, ISO 1951 has a specific element (or rather container <HeadwordCtn>) for this purpose. The corresponding rule, with a test on the 'type' attribute thus appears as follows:

```
<xsl:template match="tei:form[@type='headword']">
    <HeadwordCtn>
        <xsl:apply-templates/>
    </HeadwordCtn>
</xsl:template>
```

We also observe this case for the generic TEI element providing usage constraints, which, depending on its type attribute (e.g. 'time', 'reg') would map to specific ISO 1951 elements (resp. <TemporalUsage>, <Register>).

Conversely, whereas the TEI has a generic <gram> element for supplementary grammatical features (in our "Bahnhof" example, the provision of the appropriate determiner), ISO 1951 relies on a general purpose <Note> element to encompass such cases. The transformation rule appears in such cases as follows:

```xml
<xsl:template match="tei:gram">
    <Note type="linguisticNote">
        <xsl:apply-templates/>
    </Note>
</xsl:template>
```

The type of situations presented so far actually cover nine out of ten templates in a full TEI to ISO 1951 stylesheet. This shows how uncomplicated it is for lexicographers to derive such stylesheets, as opposed to the computational overhead that would result from the use of a more traditional programming framework. Still, there are cases where additional types of transformations are needed to cope with more in-depth discrepancies between the two models.

Just to illustrate this with one specific case, let us consider the generation of container constructs in ISO 1951 out of the generic citation objet (<cit>) in TEI. This combines two specific aspects. On the one hand, <cit> is typed when used to represent examples proper (as opposed to, for instance, translations). On the other hand, <cit> systematically contains a generic <quote> element for the embedded portion of exemplifying text, which may, or not, be accompanied by additional refinements. The mapping onto an example container (<ExampleCtn>) or a simple example element (<Example>) will depend on a specific test on the number of children that the <cit> element contains. The corresponding rules are  as follows:

```xml
<xsl:template match="tei:cit[@type='example']">
    <xsl:choose>
        <xsl:when test="child::*[2]">
            <ExampleCtn>
                <xsl:apply-templates/>
            </ExampleCtn>
        </xsl:when>
        <xsl:otherwise>
            <xsl:apply-templates/>
        </xsl:otherwise>
    </xsl:choose>
</xsl:template>
```

Furthermore, the embedded <quote> element, since it is indeed a generic quotation object in the TEI framework, has to be further tested, to ensure that it actually maps onto an
<Example> element. The corresponding rule would thus look
like:

```
<xsl:template match="tei:cit[@type='example']/tei:quote">
      <Example>
          <xsl:apply-templates/>
      </Example>
</xsl:template>
```

   As a whole, one can see that beyond the technicalities of
XSLT, the drafting of a mapping stylesheet proves to be the
best tool for the precise comparison of two models having
similar scopes, as is the case for the TEI and ISO 1951.
Moreover, it produces an actual tangible result that can be
used in several interoperability scenarios when the interface
between the two models is actually at stake.

**6. The use of schemata for consistency control (2 pp]**

We have seen so far that markup languages are an excellent
tool to separate the layout features of a text, e.g. a
dictionary article, from its logical structures. Furthermore,
with well-chosen names for elements and attributes, it is
possible to make the meaning of the respective element and
attribute values explicit to a human. With the mechanism of
assigning namespaces to elements and attributes, it is even
possible to provide, e.g. separate web pages with explicit
definitions for the elements and attributes, an additional
information source to which other developers and users can
refer. We have uses examples of namespaces in the XSLT
examples provided in section 5.2. The label "tei:", which
prefixes elements and attributes, is linked to an explanatory
document about the element and attribute examples
(xmlns:tei="http://www.tei-c.org/ns/1.0"  in the example above).

   In this context it is worth mentioning an initiative of ISO
to provide a "Data Category Registry" (cf. ISO12620:2009)
through which it is intended to provide canonical definitions
for the most widely used linguistic categories (e.g. parts of
speech, semantic cases, lexicographic field labels). In the
future, it will be possible to refer to a certain category in
this registry by prefixing an element name (e.g. "pos") with a
namespace signifier for the DCR. Everyone reading the document
can thus look up the canonical definition of the signified concept.

Document grammars like XML schemata or Relax NG schemata can also be used to control consistency during the compilation of dictionary articles. We will illustrate this with the example of the (Relax NG) schema (cf. ISO/IEC 19757-2) which is being used in the (retro-)digitization and revision of the "Wörterbuch der deutschen Gegenwartssprache". It is well known that the grammatical gender of nouns can be expressed in various ways. One can use (at least in German dictionaries) the determiner which precedes the noun in the nominative singular ('der', 'die' and 'das', respectively) or the Latin names of the genders ('Maskulinum', 'Femininum', 'Neutrum'). The latter can be abbreviated in several ways (e.g. 'f.', 'fem.' 'femin.' for the feminine gender), and this is done frequently in paper and print dictionaries, mainly to save space.

In the following, we present the content model of the "gramGrp" (= grammar group) element, specified in the Relax NG schema.

```
grammatical-description = element gramGrp {
  ...
  , (grammatical-atom-pos *
  & grammatical-atom-number ?
  & grammatical-atom-gender *
  & grammatical-atom *
  & usage-description *
  & (pronounciation-description
    | diminutive-description
    | abbreviation-description
    | variant-headword-description
    | irregular-orthography-description
  ) *
  …
  )  *
  }
```

Let us have a closer look at the "grammatical-atom-gender" item, which is highlighted in the above extract.

The abstract label "grammatical-atom-gender" is mapped to an element name "gen" which contains the gender values on an abstract as well as surface level

```
grammatical-atom-gender = element gen {
        attribute value
        { 'masculine' } , ('m.' ) )
   |(attribute value
        { 'feminine' } , ('f.' ) )
   |( attribute value
        { 'neuter' } , ('n.' ) )
}
```

The attribute values ('masculine' etc.) can later be matched to the corresponding categories of the "Data Category Registry" (see above). The schema, on the other hand, restricts the set of possible surface values to exactly one per category. Thus, the fragment presented below is valid,

```
<gramGrp>
   <pos value="N"/>
   <gen value="masculine">m.</gen>
</gramGrp>
```

while the following, slightly revised fragment is not valid.

```
<gramGrp>
   <pos value="N"/>
   <gen value="masculine">masc.</gen>
</gramGrp>
```

Of course, it is the responsibility of the dictionary producer to ensure that data is validated and unwanted abbreviations (as in the second example) are detected and corrected. Such a validator[4] can and should of course be a component of a lexicographer's workbench. The schema is only a means to enforce a consistent encoding. It is most efficient in situations where closed vocabularies – as values of attributes or as textual content of elements – can be predefined.

When it comes to deriving a representation of the data for a particular medium, either the long form, i.e. the value of the

---

[4] An example of a command based RELAG NG validator which can also be embedded in other software is jing, cf. jing:2008

attribute ('masculine'), or the abbreviated textual  content
of the element ('m') could be selected.

Schemata therefore offer a powerful tool that enables
consistency to be maintained over what are usually very large
collections of text, thereby supporting the very process of
producing lexical entries.

## 7. Conclusion

We have seen that SGML and, more importantly, XML are
adequate means for  encoding the digitized versions of paper
and print dictionaries. Entries in such dictionaries can be
modelled as trees, i.e. as a hierarchically ordered set of
functional text segments. In doing this, the sequential order
of the segments must be preserved. With XML, tree-like
structures can be modelled. Document grammars or schemata are
used to constrain the number of valid document instances.
Transformation languages like XSLT are a powerful and easy to
handle means to serialize the functional text segments in
various ways, depending on the type of media that the data
should be presented to the user in.

Furthermore, transformations can be defined to convert data
between different encoding standards.

While the TEI specification is well-known and widely used to
encode  academic dictionaries, ISO 1951 might, despite its
conceptual shortcomings, gain some influence in the world of
commercial dictionary publishing.

With the tools and formalisms we have described here, it is
possible to store lexical information in such a way that the
same data can be distributed in many ways. Moreover, the
lexical data can be visualized in many ways over a large
variety of media on many kinds of devices without ever
changing the underlying xml master data.

8. Acknowledgements

We would like to thank Julianne Nyhan for her useful comments
on earlier versions of this article.